\documentclass[10pt,twocolumn,letterpaper]{article}

\usepackage{iccv}
\usepackage{times}
\usepackage{epsfig}
\usepackage{graphicx}
\usepackage{amsmath}
\usepackage{amssymb}
\usepackage{booktabs}
\usepackage{mathtools}
\usepackage[pagebackref,breaklinks,colorlinks]{hyperref}
\usepackage{multirow}
\usepackage{makecell}
\usepackage{colortbl}
\usepackage[super]{nth}
\usepackage{subfig}
\usepackage[capitalize]{cleveref}
\crefname{section}{Sec.}{Secs.}
\Crefname{section}{Section}{Sections}
\Crefname{table}{Table}{Tables}
\crefname{table}{Tab.}{Tabs.}

\iccvfinalcopy % *** Uncomment this line for the final submission

\ificcvfinal\fi

\newcommand{\modelname}{LVDM\xspace}

\newcommand{\bm}{\mathbf{m}}
\newcommand{\bx}{\mathbf{x}}

\newcommand{\bz}{\mathbf{z}}

\newcommand{\bI}{\mathbf{I}}

\newcommand{\defeq}{\coloneqq}

\newcommand{\encoder}{\mathcal{E}}
\newcommand{\decoder}{\mathcal{D}}

\newcommand{\bzero}{\mathbf{0}}

\newcommand{\bepsilon}{{\boldsymbol{\epsilon}}}

\newcommand{\latent}{\bz}

\newcommand{\zy}[1]{\textcolor{black}{#1}}
\newcommand{\yty}[1]{\textcolor{black}{#1}}
\newcommand{\ytyn}[1]{\textcolor{black}{#1}}

\newcommand{\CUT}[1]{}

\begin{document}

\title{Latent Video Diffusion Models for High-Fidelity Long Video Generation}
\author{Yingqing He$^1$ \quad
Tianyu Yang$^2$ \quad
Yong Zhang$^2$ \quad
Ying Shan$^2$ \quad
Qifeng Chen$^1$ \\
$^1$The Hong Kong University of Science and Technology \quad
$^2$Tencent AI Lab
}
\maketitle

\ificcvfinal\thispagestyle{empty}\fi

%-------------------------------------------------------------------------

%%%%%%%%% ABSTRACT

\begin{abstract}
AI-generated content has attracted lots of attention recently, but photo-realistic video synthesis is still challenging.
Although many attempts using GANs and autoregressive models have been made in this area, the visual quality and length of generated videos are far from satisfactory.
Diffusion models have shown remarkable results recently but require significant computational resources.
To address this, we introduce lightweight video diffusion models by leveraging a low-dimensional 3D latent space, significantly outperforming previous pixel-space video diffusion models under a limited computational budget.
In addition, we propose hierarchical diffusion in the latent space such that longer videos with more than one thousand frames can be produced.
To further overcome the performance degradation issue for long video generation, we propose conditional latent perturbation and unconditional guidance that effectively mitigate the accumulated errors during the extension of video length.
Extensive experiments on small domain datasets of different categories suggest that our framework generates more realistic and longer videos than previous strong baselines.
We additionally provide an extension to large-scale text-to-video generation to demonstrate the superiority of our work.
Our code and models will be made publicly available.
\end{abstract}

%%%%%%%%% BODY TEXT

\begin{figure*}[t]
    \centering
    \includegraphics[width=1.0\linewidth]{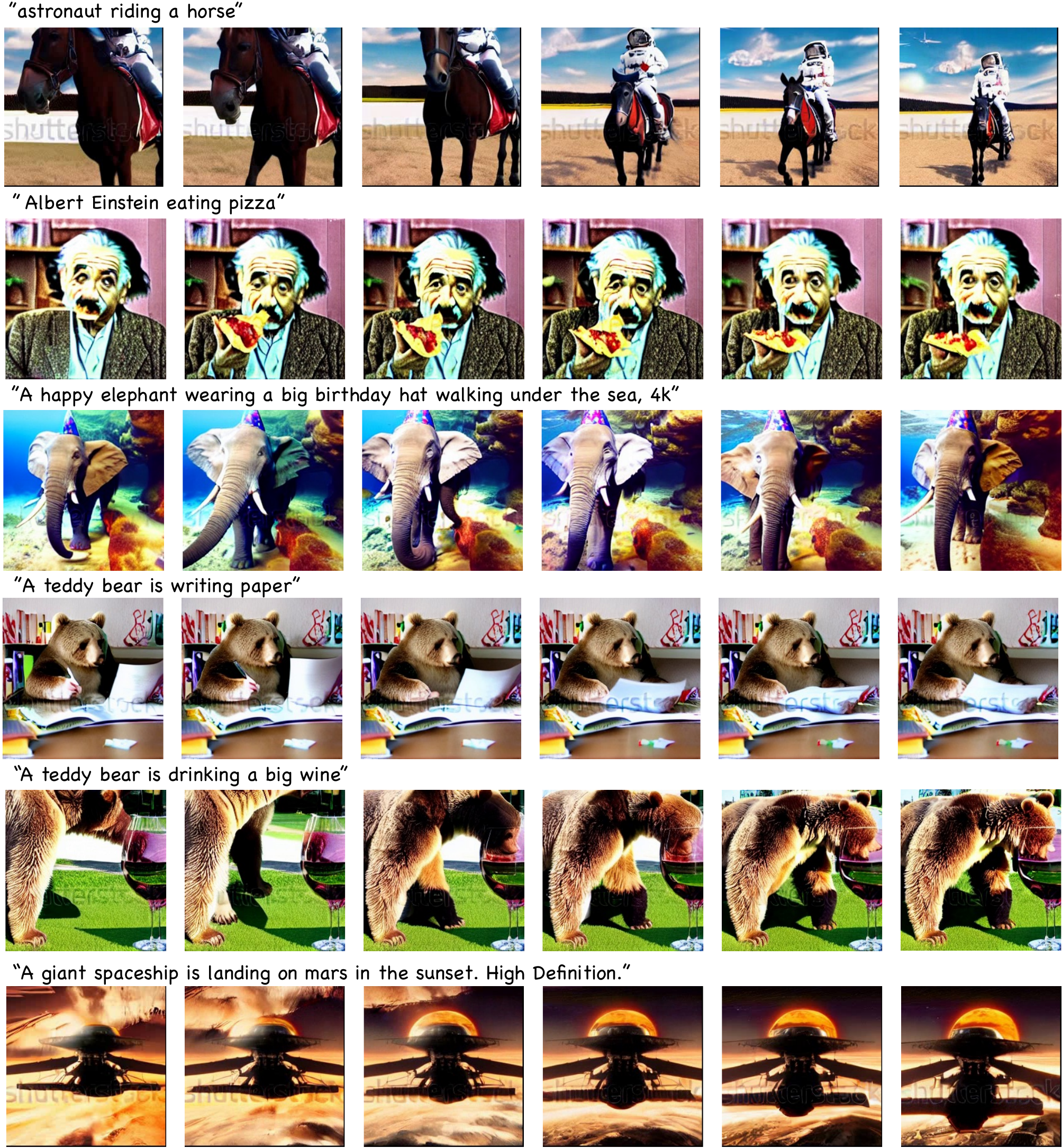}
    \vspace{-20pt}
    \caption{
    Results of extending our LVDM to text-to-video generation.
    }
    \label{fig:text2video_results}
    \vspace{-10pt}
\end{figure*}

\begin{figure*}[t]
    \centering
    \includegraphics[width=1.0\linewidth]{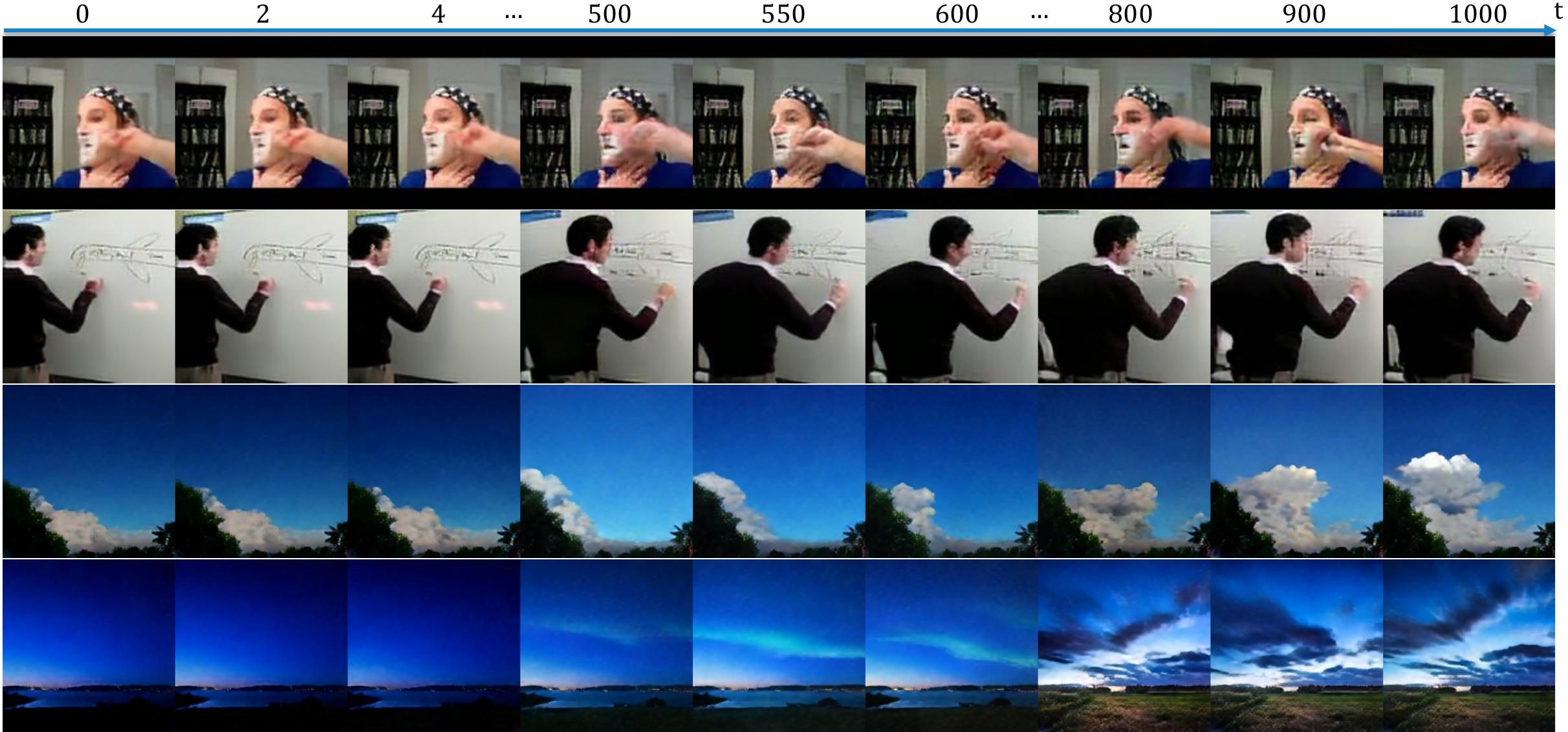}
    \vspace{-20pt}
    \caption{
        Our unconditional long video generation results on UCF-101~\cite{UCF101} and Sky Time-lapse~\cite{SkyTimelapse_dataset}.
        $t$ indicates the frame index.
    }
    \label{fig:teaser}
    \vspace{-10pt}
\end{figure*}

\section{Introduction}
\label{sec:intro}

A video can provide more informative, attractive, and immersive visual content that presents the physical 3D world. A powerful video generation tool can benefit novel content creation, gaming, and movie production.
Thus, rendering photorealistic videos is a long-standing and exciting goal of the computer vision and graphics research community.
However, the high-dimensional video samples and the statistical complexity of real-world video distributions make video synthesis quite challenging and computationally expensive.
%
% The synthesizing quality is also far from satisfactory, which limits its real-world applications.

Existing works capitalize on different types of generative models including GANs~\cite{vgan, tgan, tgan-v2, mocogan, mocogan-hd, Stylegan-V, long-video-gan}, VAEs~\cite{video-vae, VideoGPT}, autoregressive models~\cite{tats, scale-autoregressive, latent-video-transformer}, and normalizing flows~\cite{videoflow}.
Particularly, GANs have achieved great success in image generation~\cite{biggan, stylegan, stylegan2, stylegan3}, thus extending it to video generation with a dedicated temporal design achieves outstanding results~\cite{mocogan-hd, Stylegan-V}.
However, GANs suffer from mode collapse and training instability problems, which makes GAN-based approaches hard to scale up to handle complex and diverse video distributions.
%
% Besides, many GAN-based methods usually assume that different frames share similar content in a video, which limits the ability to produce videos with novel content emerging over time\cite{mocogan-hd,Stylegan-V}.
%
% Recent GAN-based methods attempt to exploit implicit neural representations of GANs\cite{digan,Stylegan-V} to model motion as continuous representation, which is hard to generate long videos of more than 1000 frames.
% encounter the problem of periodic motion and performance degradation across time.
%
Most recently, TATS~\cite{tats} proposes an autoregressive approach that leverages the VQGAN~\cite{vq-gan} and transformers to synthesize long videos.
However, the generation fidelity and resolution (128$\times$128) still have much room for improvement.

To overcome these limitations, we leverage diffusion models (DMs)~\cite{ddpm}, another class of generative models that achieve impressive performance in various image synthesis tasks~\cite{adm, dalle-2, glide, imagen, ldm}.
However, directly extending DMs to video synthesis requires substantial computational resources~\cite{video-dm, mcvd}.
%
% Thus, we aim to leverage the powerful representation ability of diffusion models while maintaining a reasonable computing budget for video synthesis.
%
Besides, most of the text-to-video generation models are not available to the public~\cite{imagen-video, make-a-video, magicvideo, video-dm}, which hinders the research progress of this field.
To tackle these problems, we devise LVDM, an efficient video diffusion model in the latent space of videos and we achieve state-of-the-art results via the simple base LVDM model.
% We achieve this by projecting videos to the latent space via a 3D autoencoder and performing denoising and diffusion processes in this low-dimensional space.
%
In addition, to further generate long-range videos, we introduce a hierarchical LVDM framework that can extend videos far beyond the training length.
However, generating long videos tends to suffer the performance degradation problem.
To mitigate this issue, we propose conditional latent perturbation \ytyn{and unconditional guidance}, which effectively slow the performance degradation over time.
Ultimately, our framework surpasses many previous works in short and long video generation and establishes new state-of-the-art performance on three datasets. We also provide additional results for text-to-video generation as an extension.
Notably, our codes and pre-trained models will be publicly available as an efficient diffusion baseline model for video synthesis and downstream video editing tasks.
% on multiple video generation benchmarks.

In sum, our work makes the following contributions:
\begin{itemize}
    \item We introduce \modelname, an efficient diffusion-based baseline approach for video generation by firstly compressing videos into tight latents.
 % along spatial and temporal axis.
 % the first latent-based video diffusion model.
    \item We propose a hierarchical framework that operates in the video latent space, enabling our models to generate longer videos beyond the training length further.
    % , enabling our models to generate longer videos beyond the training lengths (more than 1,000 frames).
	% \item We discover that the performance degradation issue can be effectively mitigated by employing \textit{conditional latent perturbation} and \textit{unconditional guidance} techniques during long video generation.
    \item We propose \textit{conditional latent perturbation} and \textit{unconditional guidance} for mitigating the performance degradation issue during long video generation.
    \item Our model achieves state-of-the-art results on three benchmarks in both short and long video generation settings. We also provide appealing results for open-domain text-to-video generation, demonstrating the effectiveness and generalization of our models.
    % \item Our code and models will release to the public.
\end{itemize}
% \clearpage

\section{Related Work}
\label{sec:related_work}

\begin{figure*}[t]
    \centering
    \includegraphics[width=1.0\linewidth]{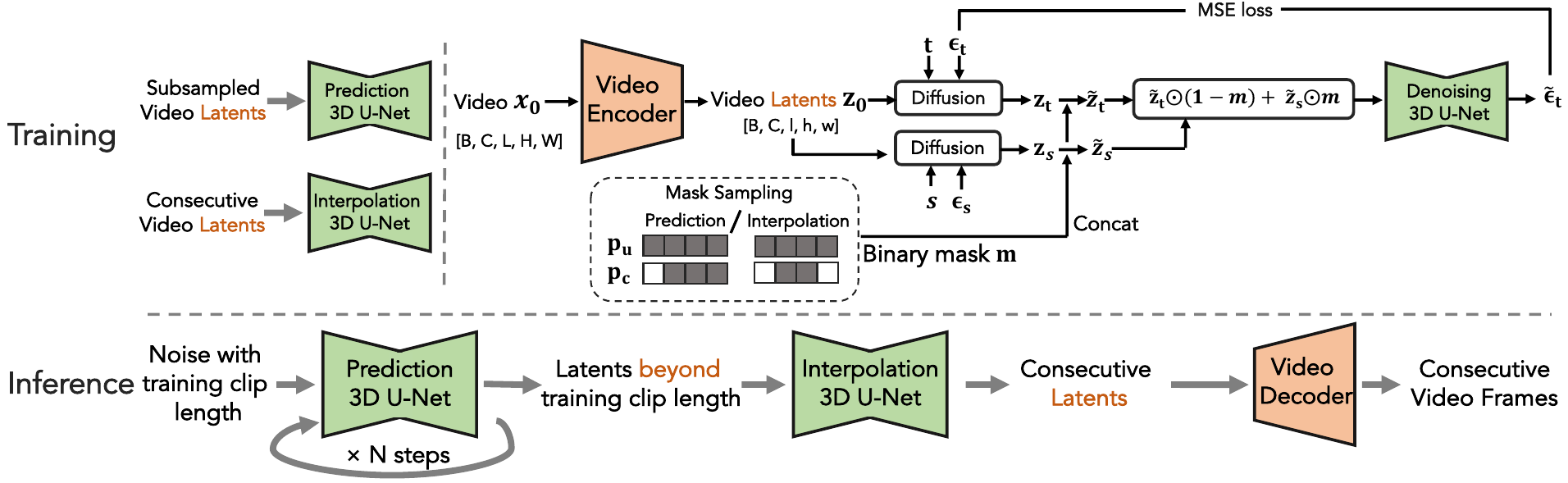}
    \vspace{-15pt}
    \caption{
    \textbf{Hierarchical LVDM Framework.}
    We present a hierarchical latent video diffusion model (LVDM) for generating longer videos beyond the temporal training length.
    $t$ and $s$ are randomly sampled diffusion timesteps for generated latents and conditional latents, respectively.
    $p_c$ and $p_u$ are probabilities of the conditional and unconditional input, respectively.
    % \modelname, a novel diffusion model (DM)-based framework for video generation.
    %
    % After a video autoencoder is trained, we obtained video latents of consecutive frames.
    %
    % Then we train two conditional LVDMs one for autoregressively extending the generated video length, and another one for interpolating middle latents to increase the video frame rate.
    %
    % We subsampled the video latents instead of subsampling the video frames to reuse the autoencoder trained on consecutive frames.
    %
    % The diffusion and denoising process is performed on the video latent space, which is learned by a \textit{3D autoencoder}.
    %
    % Then an \textit{unconditional DM} is trained on the latent space for generating short video clips.
    %
    % To extend videos to arbitrary lengths, we further propose two frame-conditional models, including a \textit{prediction DM} and an \textit{infilling DM} which can synthesize long-duration videos in autoregressive and hierarchical ways.
    %
    % We utilize noisy conditions at diffusion timestep $s$ to mitigate the condition error induced during the autoregressive sampling process.
    %
    % The frame-conditional DMs are jointly trained with unconditional inputs, where the conditional and unconditional sample frequencies are controlled by their corresponding probabilities, i.e., $p_c$ and $p_u$.
    }
    \label{fig:framework}
    \vspace{-10pt}
\end{figure*}

\subsection{Video Synthesis}
Video synthesis aims to model the distribution of real-world videos, and then one can randomly draw realistic and novel video samples from the learned distribution.
% In this work, we mainly focus on the unconditional video generation setting.
% TODO: video prediction, interpolation, unconditional genration(from noise) is the most diffucult. 
Prior works mainly exploit deep generative models, including GANs~\cite{digan,Stylegan-V, mocogan-hd, mocogan,tgan-v2,tgan, vgan}, autoregressive models~\cite{VideoGPT,tats}, VAEs~\cite{vqvae-prediction,video-vae}, and normalizing flows~\cite{videoflow}. 
Among them, the most dominant ones % in the past few years 
are GAN-based approaches due to the great success of GANs in image modeling.
MoCoGAN~\cite{mocogan} and MoCoGAN-HD~\cite{mocogan-hd} learn to decompose latent codes into two subspaces, i.e., content and motion.
MoCoGAN-HD~\cite{mocogan-hd} leverages the powerful pre-trained StyleGAN2 as the content generator, demonstrating higher-resolution video generation results.
StyleGAN-V~\cite{Stylegan-V} and DiGAN~\cite{digan} introduce implicit neural representation to GANs for modeling the continuity of temporal dynamics. 
They built long-video GAN on top of StyleGAN3 and apply hierarchical generator architecture for long-range modeling, thus producing videos with new content arising in time.
Despite these achievements made by GANs, those methods tend to suffer from mode collapse and training instability.
Autoregressive methods have also been exploited for video generation.
VideoGPT~\cite{VideoGPT} uses VQVAE~\cite{VQ_VAE} and transformer to autoregressively generate tokens in a discrete latent space.
TATS~\cite{tats} changes the VQVAE~\cite{VQ_VAE} to a more powerful VQGAN~\cite{vq-gan} and combines a frame interpolation transformer to render long videos in a hierarchical manner.
Different from the aforementioned methods, we study diffusion models for video generation in this work.
%high-fidelity and arbitrary long 

\subsection{Diffusion Models}
Diffusion models are a class of likelihood-based generative models that have shown remarkable progress in image synthesis tasks.
Due to their desirable properties like stable training and easy scalability, %advantages including the high-dimensional latent space \NOTE{why advantages}, simple training objective, and progressive denoising procedure, 
diffusion models have surpassed GANs on the image generation task and achieved both higher sample quality and better distribution coverage~\cite{adm}.
Its scalability further facilitates the building of large-scale text-to-image generation models~\cite{imagen,glide,ldm,dalle-2}, which show marvelous image samples recently.
The groundbreaking work of diffusion models is DDPM~\cite{ddpm}, which leverages the connection between Langevin dynamics~\cite{dynamic} and denoising score matching~\cite{score-sde} to build a weighted variational bound for optimization.
% Langevin dynamics:Deep unsupervised learning using nonequilibrium thermodynamics.
% denoising score matching: Generative Modeling by Estimating Gradients of the Data Distribution nips 2019 
However, the sampling process needs to follow the Markov chain step by step to produce one sample, which is extremely slow (e.g., usually 1000 or 4000 steps~\cite{ddpm,adm}).
DDIM~\cite{ddim} accelerates the sampling process via an iterative non-Markovian way while keeping the same training process unchanged.
~\cite{im-ddpm} further improves the log-likelihoods while maintaining high sample quality, and ADM~\cite{adm} eventually outperforms GAN-based methods via an elaborately designed architecture and classifier guidance.
% , i.e., an extra noisy image classifier to guide the class-conditional generation while improve sample fidelity.
% , which is used to boost the sample fidelity for class-conditional generation.
%
Contemporarily, a cascaded diffusion model~\cite{cascaded-dm} is proposed with a sequence of lower resolution conditional diffusion models as an alternative approach to improve the sample fidelity and shows that conditional noisy augmentation is pivotal to the stacked architecture.
Despite their findings and achievement in image synthesis, diffusion models on video generation have not been well-studied.
Most recently, VDM~\cite{video-dm} extends diffusion models to the video domain, which initiates the exploration of diffusion models on video generation.
Specifically, they modify the 2D UNet to a spatial-temporal factorized 3D network and further present image-video joint training and gradient-based video extension techniques.
MCVD~\cite{mcvd} parameterizes unconditional video generation and conditional frame prediction and interpolation models as a unified one by randomly dropping conditions (previous or future frames) during training, which shares the similar underneath idea with classifier-free guidance which joint trains a class-conditional and unconditional model.
Make-A-Video~\cite{make-a-video}, and Imagen Video~\cite{imagen-video} leverage a series of big diffusion models for large-scale video synthesis conditioned on a given text prompt.
However, previous video-based video generation approaches all perform diffusion and denoising processes in pixel space, which requires substantial computational resources.
%
%To our best knowledge, we are the first one to explore efficient video diffusion models via generation on the low-dimentional video latent space.
%
In this paper, we extend the latent image diffusion model~\cite{ldm} to video by devising a 3D auto-encoder for video compression.
Founded on this baseline, we further show how long videos can be sampled via a hierarchical architecture and natural extensions of conditional noise augmentation.
% and classifier-free guidance to video generation.

\textbf{Concurrent works.}
With the rapid development of diffusion models, two contemporary works share similar ideas with us and propose latent-based diffusion models.
MagicVideo~\cite{magicvideo} proposes an efficient video generation framework via latent diffusion models.
Unlike MagicVideo, which compresses videos frame-by-frame, we compress the redundant information along the temporal axis to obtain more compact latent.
PVDM~\cite{pvdm} proposes to parameterize videos as image-like latent by selecting three 2D latents along three axes of the 3D video latent. Then they train a 2D diffusion network to model these 2D latents for video. Differently, we exploit 3D diffusion networks to model the cubic video latents directly.
Another important difference between our work with Magic video and PVDM is that we make a further step towards long video generation and provide a hierarchical framework, conditional noise perturbation and unconditional guidance to boost the long video generation performance.
on alleviating the performance degradation problem.

\section{Method}
\label{sec:method}

% \cref{fig:framework} shows our overall framework.
We firstly compress video samples to a lower-dimensional latent space by a video autoencoder.
Then we design a unified video diffusion model, which can perform both unconditional generation and conditional video generation in one network, in the latent space. This enables our model to self-extend the generated video to an arbitrary length autoregressively.
However, autoregressive models only tend to suffer the problem of performance degradation over time.
To further improve the coherence of generated long video and alleviate the quality degradation problem, we propose hierarchical latent video diffusion models to first generate video patents sparsely and then interpolate intermediate latents.
We also propose conditional latent perturbation and unconditional guidance for tackling the performance degradation problem in long video generation.
% using an autoregressive model and then interpolate it to a higher frame rate by an interpolation diffusion model.
% \yty{Note that both the autoregressive and interpolation diffusion model are trained in a similar way, except the input clips have different strides.}
%Specifically, the latent video diffusion model can perform three video synthesis tasks: unconditional sampling, video prediction, and interpolation, in one network.
%
%This makes our methods can increase the frame rate via latent interpolation, and self-extend the generated video to an arbitrary length by latent prediction.
%

% \subsection{Video Compression via a 3D Autoencoder}
\subsection{Video Autoencoder}

\label{subsec:compress}
% back+problem+thus
%One of the reasons for the high-fidelity sampling ability of diffusion models is the large latent space, which is a series of latents across time $t$ ($t \sim \mathrm{Uniform}(\{1, \dotsc, T\})$), and each latent in time $t$ has the same dimension with data samples.
%
%Directly extending image DMs to video DMs causes a substantial computational cost.
%
%More specifically, consider a short video clip of shape 256 $\times$256$\times$16, the latent dimension will be one million, which is  16 times of image diffusion models of the same spatial resolution, and is 2048 times of image GAN which is the learned $w$ space~\cite{stylegan2} usually has the latent dimension of 512.
%
%Training and sampling on this million-level latent space not only require huge computational resources but also slow down the training and sampling speed.
%
%Thus, to reduce the computational cost while exploiting diffusion models' modeling capacity, we propose to apply diffusion models on a more compact video latent space.

We compress videos using a lightweight 3D autoencoder, including an encoder $\encoder$ and a decoder $\decoder$. Both of them consist of several layers of 3D convolutions.
Formally, given \zy{a video sample} $\bx_0 \sim p_{data}(\bx_0)$ where $\bx_0 \in \mathbb{R}^{H \times W \times L \times 3}$, the encoder $\encoder$ encodes it to its latent \zy{representation} $\bz_0=\encoder(\bx_0)$ where $\bz_0 \in \mathbb{R}^{h \times w \times l \times c}$, $h=H/f_s, w=W/f_s$, \zy{and} $l=L/f_t$. $f_s$ and $f_t$ are spatial and temporal downsampling factors.
The decoder $\decoder$ decodes $\bz_0$ to the reconstructed \zy{sample} $\tilde{\bx}_0$, \yty{i.e.} $\tilde{\bx}_0 = \decoder(\bz_0)$.
To ensure \yty{that} the autoencoder is \yty{temporally shift-equivariant}, we follow~\cite{tats} to use repeat padding in all three-dimensional convolutions.
 % kernels
The training objective includes \zy{a} reconstruction loss $\mathcal{L}_{rec}$ and \zy{an} adversarial loss $\mathcal{L}_{adv}$.
The reconstruction loss $\mathcal{L}_{rec}$  \zy{is comprised of} a pixel-level mean-squared error (MSE) loss %$\mathcal{L}_{mse}$ 
and a perceptual-level LPIPS~\cite{lpips} loss. %$\mathcal{L}_{lpips}$.
The adversarial loss~\cite{vq-gan} is used to eliminate the \zy{blur in} reconstruction usually caused by the pixel-level reconstruction loss and further improve the \zy{realism of the reconstruction}.
In summary, the overall training objective of $\encoder$ and $\decoder$ is
\yty{\begin{equation}
		\begin{aligned}
	\mathcal{L}_{AE} = \min_{\encoder,\decoder}\max_{\psi}\bigl (\mathcal{L}_{rec}(\bx_0, \decoder(\encoder(\bx_0)))  \\+ \mathcal{L}_{adv}(\psi(\decoder(\encoder(\bx_0)))\bigr ).
	% \mathcal{L}_{mse}=\Vert \bx_0 - \hat{\bx_0} \Vert^2; \\
	% \mathcal{L}_{adv} = \log D(x) + \log (1 - D(\hat{x}))
\end{aligned}
\end{equation}
where $\psi$ is the discriminator used in adversarial training.}

% ---------------------------------------------------------------------------------------
% \input{figures/scale_factor}
% \subsection{Short Video Generation}
\subsection{Base LVDM for Short Video Generation}
\noindent\yty{\textbf{Revisiting Diffusion Models.}} We propose \yty{to perform} diffusion and denoising on the video latent space.
Given \zy{a} compressed latent code $\bz_0 \sim p_{data}(\bz_0)$, we train diffusion models to generate latent samples \zy{starting} from a pure Gaussian noise $\bz_T \sim \mathcal{N}(\bz_T; \bzero, \bI)$ in $T$ timesteps, \zy{producing a set of noisy latent variables, i.e.,} {$\bz_1$, ..., $\bz_T$} .
% t ~ Uniform({1, ..., T})
The forward diffusion process is gradually adding noise to $\bz_0$ according to a predefined variance schedule $\beta_1, \dotsc, \beta_T$:
\begin{align}
q(\latent_{1:T} | \latent_0) &\coloneqq \prod_{t=1}^T q(\latent_t | \latent_{t-1} ),\\
q(\latent_t|\latent_{t-1}) & \coloneqq \mathcal{N}(\latent_t;\sqrt{1-\beta_t}\latent_{t-1},\beta_t \bI). \label{eq:forwardprocess}
\end{align}

% \end{align}

Eventually, the data point $\latent_T$ becomes indistinguishable from pure Gaussian noise.
To recover $\latent_0$ from $\latent_T$, diffusion models learn a backward process via
\begin{align}
p_\theta(\latent_{0:T}) & \coloneqq p(\latent_T) \prod_{t=1}^{T} p_\theta(\latent_{t-1}|\latent_t), \\
p_\theta(\latent_{t-1}|\latent_t) & \coloneqq \mathcal{N}(\latent_{t_1}; \mu_\theta(\latent_t, t), \Sigma_\theta(\latent_t, t)),
\label{eq:backwardprocess}
\end{align}
where $\theta$ is a parameterized neural network, typically a U-Net~\cite{unet} commonly used in image synthesis, to predict $\mu_\theta(\latent_t, t)$, and $\Sigma_\theta(\latent_t, t)$. \yty{In practice, we parameterize $\mu_\theta(\latent_t, t)$ by
\begin{equation}
	\mu_\theta(\latent_t, t) = \frac{1}{\sqrt{\alpha_t}}\left (\latent_t-\frac{\beta_t}{\sqrt{1-\bar{\alpha}_t}} \epsilon_\theta(\latent_t, t) \right),
\end{equation}
where $\epsilon_\theta(\latent_t, t)$ is eventually estimated.
% , which is shown to work best~\cite{ddpm}. 
We simply set $\Sigma_\theta(\latent_t, t)=\beta^2_t\bI$ as in~\cite{ddpm}}.
% or directly predict $\epsilon_t$

The training objective is a \yty{simplified} version of variational bound:
% \begin{align}
%  L_\mathrm{simple}(\theta) \defeq \Eb{t, \bx_0, \bepsilon}{ \left\| \bepsilon - \bepsilon_\theta(\sqrt{\bar\alpha_t} \bx_0 + \sqrt{1-\bar\alpha_t}\bepsilon, t) \right\|^2} \label{eq:training_objective_simple}
% \end{align}
\begin{align}
 \mathcal{L}_\mathrm{simple}(\theta) \defeq \left\| \bepsilon_\theta(\latent_t, t) - \bepsilon \right\|_2^2 \label{eq:training_objective_simple},
\end{align}
where $\bepsilon$ is drawn from a diagonal Gaussian distribution.
% \todo{transition text to estimate $\bepsilon$}

\noindent\yty{\textbf{Video Generation Backbone.}} To model video samples in the 3D latent space, we follow~\cite{video-dm} that exploits a \yty{spatial-temporal factorized 3D U-Net} architecture to estimate the $\bepsilon$.
Specifically, we use \yty{space-only 3D} convolution with the shape of $1 \times 3 \times 3$ and add \yty{temporal} attention in partial layers.
We \zy{investigate} two kinds of attention: \yty{joint spatial-temporal} self-attention and \yty{factorized} spatial-temporal self-attention.
We \yty{observe that applying the joint spatial-temporal attention does not exhibit  significant benefit compared with the factorized one while increasing the model complexity and introducing spot-like artifacts in random locations sometimes. }% (See \cref{fig:scale_factor_effect}).} %3D attention has no significant benefit to the generation performance and contrarily introduce spot-like artifact in random locations.
Thus we use factorized \yty{spatial-temporal attention as the default setting} in our experiments.
We use adaptive group normalization to inject the timestep embedding into normalization modules to control the channel-wise scale and bias parameters, which have also \yty{been} demonstrated \yty{to be beneficial} for improving sample fidelity in~\cite{adm}.
% For better timestep information \yty{injection}, w

% \subsection{Conditional Latent Prediction and Interpolation}
% \subsection{Long Video Generation}
%\subsection{Hierarchical Latent Diffusion Models}
\subsection{\ytyn{Hierarchical LVDM for Long Video Generation}}
The aforementioned framework can only generate short videos, whose lengths are determined by the input frame number during training.
%\todo{long video eq.}To extend sampled video length, 
We therefore propose a conditional latent diffusion model, which can produce future latent codes conditioned on the previous ones in an autoregressive manner, to facilitate long video generation. \ytyn{We further present several techniques to alleviate the error accumulation problem in autoregressive generation.}% Fig. \ref{fig:framework} illustrates the whole framework.}

\noindent\textbf{Autoregressive Latent Prediction.} 
Considering a short clip latent $\bz_t = \{\bz_t^i\}_{i=i}^l$ where $\bz_t^i \in \mathbb{R}^{h \times w \times c} $ \zy{and} $l$ is the number of latent codes within the clip, we can learn to predict future latent codes conditioned on the former ones. For each video frame in a clip latent, we add an additional binary mask along the channel dimension to indicate whether it is a conditional frame or a frame \zy{to predict} \ytyn{and replace the $\bz_t^i$ with $\bz_0^i$ according to the mask, yielding, 
\begin{equation}
\begin{split}
    \tilde{\bz}_t &= \{\tilde{\bz}_t^i=[\bz_t^i,\bm^i]\}_{i=1}^l\}_{i=1}^l, \tilde{\bz}_t^i \in \mathbb{R}^{h \times w \times (c+1)} \\
    \tilde{\bz}_0 &= \{\tilde{\bz}_0^i=[\bz_0^i,\bm^i]\}_{i=1}^l\}_{i=1}^l, \tilde{\bz}_0^i \in \mathbb{R}^{h \times w \times (c+1)}\\
    \tilde{\bz}_t &\leftarrow \tilde{\bz}_t\odot(1-\bm) + \tilde{\bz}_0\odot\bm
\end{split}
\end{equation}
where $\bm = \{\bm^i\}_{i=1}^l, \bm^i \in \mathbb{R}^{h \times w \times 1}$ is the binary mask.}
%Formally, given \zy{a} binary mask clip $\bm = \{\bm^i\}_{i=1}^l$ where $\bm^i \in \mathbb{R}^{h \times w \times 1}$, we obtain the conditional input clip as $\tilde{\bz}_t = \{\tilde{\bz}_t^i=[\bz_t^i,\bm^i]\}_{i=1}^l$ where $\tilde{\bz}_t^i \in \mathbb{R}^{h \times w \times (c+1)}$. 
By randomly setting different binary masks to ones or zeros, we can train our diffusion model to perform both unconditional video generation and conditional video prediction jointly. 
Concretely, we set all masks in the binary clip $\bm$ to \zy{zeros} for unconditional diffusion model training. %For conditional video prediction, 
\ytyn{During inference stage,} we set the first $k$ binary mask $\{\bm^i\}_{i=1}^{k}$ to \zy{ones} and the remaining  $\{\bm^i\}_{i=k+1}^{l}$ to \zy{zeros}. %At the same time, we replace the first $k$ latent codes with the ground truth, \ie $\{\bz_t^i\}_{i=1}^{k}\leftarrow\{\bz_0^i\}_{i=1}^{k}$}

\noindent\yty{\textbf{Hierarchical \ytyn{Latent} Generation.}} \yty{Generating videos in an autoregressive way has the risk of quality degradation caused by accumulated error over time. We thus utilize a common strategy, hierarchical generation~\cite{tats, cogvideo, hierar_video_gen}, to alleviate this problem. Specifically, we first train an autoregressive video generation model on \emph{sparse frames} to form the basic storyline of the video and then train another interpolation model to \ytyn{infill} the missing frames. The training of the interpolation model is similar to the autoregressive model, except that we set the binary masks of \zy{the} middle frames between \zy{every two} sparse frames to \zy{zeros}. }

% \CUT{we can select the past latents\todo{} $\bz_{past} \in \mathbb{R}^{h \times w \times L_{past} \times c}$ as condition and continuously predict future latents based on the past ones.
% %
% One can also train a latent interpolation network to increase the frame rate of $\bz$.
% % 
% % A combination of \uncondnet and \intepnet can also gen longer videos in a hie way.
% To increase the video length in a small number of prediction steps to slow down the performance degradation problem~\cite{tats}, one can train an unconditional generation \modelname with a latent frame stride $fs_l$ to produce long but sparse latent variables, then with an extra latent interpolation network, one can increase the latent frame rate to be the same as the original latent speed.
% % make model condition-aware
% To emphasize the condition information across time, we further design a binary map as an indicator to specify the condition location, where all zero masks indicate the unconditional generation and ones indicate the conditional positions.
% %
% We further unify three types of video generation tasks through joint training.
% %
% During training, the mask is randomly generated with three specified probabilities: $p_u$, $p_p$, and $p_i$, where $p_u + p_p + p_i = 1$.
% %
% In practice, we first train an unconditional model and then resume it with joint training.}

\noindent\textbf{Conditional Latent Perturbation.}
Although the above-mentioned hierarchical generation manner can reduce the number of autoregressive steps to overcome the degradation issue,
% has demonstrated its effectiveness in prior works~\cite{tats, cogvideo}, 
more prediction steps are indispensable to produce long-enough video samples.
Thus, we propose conditional perturbation to mitigate the conditional error induced by the previous generation step.
Specifically, rather than directly \yty{conditioning on $\latent_0$}\CUT{condition frames at t=0}, we use the noisy latent code \yty{$\latent_s$} at an arbitrary time $s$, \yty{which could be computed by \eqref{eq:forwardprocess}, as the condition} during training, \yty{\ie, $\{\bz_t^i\}_{i=1}^{k}\leftarrow\{\bz_s^i\}_{i=1}^{k}$.} This means we also perform a diffusion process on the conditional frames.
% noise augmentation formula.
% corrupt the condition information to reduce the performance degradation has also been applied in the NLP area. 
%
To keep the conditional information preserved, a maximum threshold $s_{max}$ is used to clamp the timesteps in a minor noise level.
During sampling, a fixed noise level is used to consistently add noise during autoregressive prediction.
Conditional latent perturbation \zy{is} inspired by conditional noise augmentation, which has been proposed in cascaded diffusion models~\cite{cascaded-dm} to improve the performance of super-resolution diffusion models.
%as augmentation techniques.
%
% While we extend it to video prediction, we are the first to demonstrate its effectiveness \zy{in} producing long video samples.
We extend it to video prediction, and we are the first to demonstrate its effectiveness \zy{in} producing long video samples.
%We further design two implementation variants of injecting the timestep s to the network.\todo{a fig or equation}
% the conditional noise purturbation:.

\noindent\textbf{Unconditional Guidance.}
Another complementary \yty{technique to alleviate quality degradation of autoregressive video generation} is to leverage the unconditional score to guide the conditional \yty{generation process.}\CUT{task for reducing performance degradation.} \yty{Since the accumulated error during autoregressive generation does not affect the unconditional score, introducing this score into long video generation could improve the diversity and fidelity of sampled video.}
Thanks to the joint training techniques presented above, we can use one network to estimate both unconditional scores $\bepsilon_u$ and conditional scores $\bepsilon_c$. \yty{By zeroing all binary maps $\{\bm^i\}_{i=1}^l$ in $\tilde{\bz}$, we obtain $\bepsilon_u$. By setting the first $k$ binary maps $\{\bm^i\}_{i=1}^{k}$ to one and the remaining ones $\{\bm^i\}_{i=k+1}^{l}$ to zero in $\tilde{\bz}$, we get $\bepsilon_c$.}
%\CUT{\begin{align}
%\bepsilon_u = \bepsilon_\theta(\bz_t); \qquad \bepsilon_c = \bepsilon_\theta(\bz_t, \bc)
%\end{align}}
%
Note that the conditional score may be out of the model learned distribution due to the error accumulation when autoregressively predicted.
Thus we propose to leverage the unconditional scores to guide the prediction sampling process via
\begin{align}
    \tilde{\epsilon}_\theta%(\bz_t, \bc) 
    = (1+w)\bepsilon_c - w\bepsilon_u, 
\label{eq:classifier_free_score}
\end{align}
where $w$ is the guidance strength.
\CUT{\textbf{Discussions.} }
This formula is initially presented in~\cite{classifier-free} and referred to as \textit{classifier-free guidance} to avoid training a separate classifier for class-conditional diffusion models.
% %
% It also has the ability to adjust the trade-off between sample quality and distribution coverage (like the truncation in GANs) and has shown performance gains in text-conditional image synthesis~\cite{glide, ldm}.
% %
We extend this idea to guide the frame-conditional diffusion models to generate longer videos.

\section{Experiments}

\begin{figure*}[t]
    \centering
    \includegraphics[width=1.0\linewidth]{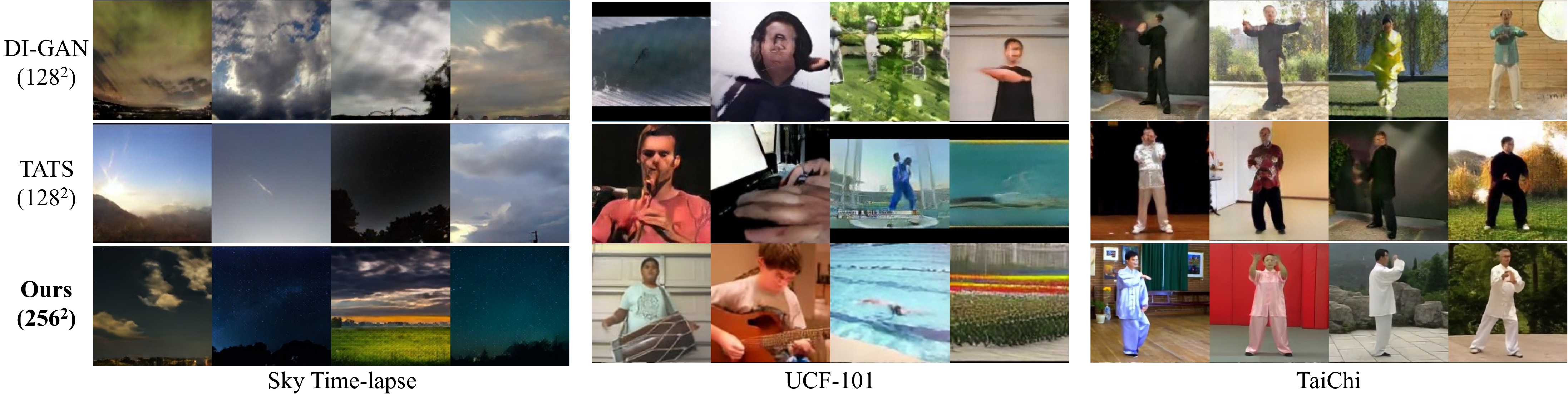}
    \vspace{-20pt}
    \caption{
    Qualitative comparison with state-of-the-art methods on unconditional short video generation (16 frames) on three datasets.
    % on UCF-101\cite{UCF101} Sky Time-lapse\cite{SkyTimelapse_dataset}, and TaiChi\cite{taichi}.
    %
    % We showcase results including  DIGAN\cite{digan}, TATS\cite{tats}, and \modelname (Ours).
    %
    % \modelname can generate videos with 16 frames simultaneously under 256$^2$ resolution and achieves higher resolution while better quality results (Zoom in for better view).
    %due to the compact latent space diffusion 
    % Samples of baseline methods are obtained from their official checkpoints.
    %
    All frames shown are selected at $t=10$ from four \textit{randomly} generated samples.
    }
    \label{fig:short_video_quali_compare}
    \vspace{-0pt}
\end{figure*}

\begin{table*}[t] 
    \centering
    \subfloat[
    Sky Time-lapse
    \label{tab:sky-short}
    ]{
    \begin{minipage}[l]{.32\linewidth}
    \centering
    \resizebox{1.0\linewidth}{!}{
        \begin{tabular}{lcc}
            \toprule
            Method & FVD$_{16}\downarrow$ & KVD$_{16}\downarrow$ \\
            \toprule
            \multicolumn{3}{l}{\textit{Resolution 128$^2$}} \\
            \midrule
            MoCoGAN-HD~\cite{mocogan-hd} & 183.6 $\pm$ 5.2 & 13.9 $\pm$ 0.7 \\
            DIGAN~\cite{digan} & 114.6 $\pm$ 4.9 & 6.8 $\pm$ 0.5 \\
            TATS~\cite{tats} & 132.6 $\pm$ 2.6 & 5.7 $\pm$ 0.3 \\
            Long-video GAN~\cite{tats} & 107.5 & - \\
            \midrule
            \multicolumn{3}{l}{\textit{Resolution 256$^2$}} \\
            \midrule
            Long-video GAN~\cite{tats} & 116.5 & - \\
            % \midrule
            \textbf{\modelname (Ours)} & \textbf{95.2 $\pm$ 2.3} & \textbf{3.9 $\pm$ 0.1}\\
            \bottomrule
        \end{tabular}
        }
    \end{minipage}
    }
    \hspace{4pt}
    \subfloat[
    UCF-101
    \label{tab:short-ucf}
    ]{
    \begin{minipage}[c]{.30\linewidth}
        \centering
        \resizebox{1\linewidth}{!}{
        \begin{tabular}{lcc}
            \toprule
            Method & FVD$_{16}\downarrow$ & KVD$_{16}\downarrow$ \\
            \toprule
            \multicolumn{3}{l}{\textit{Resolution 64$^2$}} \\
            \midrule
            MCVD~\cite{mcvd} & 1143 & -  \\
            \midrule
            \multicolumn{3}{l}{\textit{Resolution 128$^2$}} \\
            \midrule
            TGAN-v2~\cite{tgan-v2} & 1209 $\pm$ 28 & - \\
            % DIGAN~\cite{digan} & 655 $\pm$ 22 & 29.71 $\pm$ .53 & - \\
            DIGAN$^*$~\cite{digan} & 577 $\pm$ 21 & - \\
            TATS~\cite{tats} & 420 $\pm$ 18 & - \\
            % \textbf{\modelname (Ours, resizing)} & 429 $\pm$ 20 & & 32.21 $\pm$ 2.52 \\
            \midrule
            \multicolumn{3}{l}{\textit{Resolution 256$^2$}} \\
            \midrule
            MoCoGAN-HD$^*$~\cite{mocogan-hd} & 700 $\pm$ 24 &  - \\
            % TATS~\cite{tats} & 635 $\pm$ 34 & 55 $\pm$ 5 \\
            \textbf{\modelname (Ours$^*$)} & \textbf{372 $\pm$ 11} & \textbf{27 $\pm$ 1} \\
            \bottomrule
        \end{tabular}
        }
    \end{minipage}
    }
    \hspace{4pt}
    \subfloat[
    Taichi
    \label{tab:short-taichi}
    ]{
    \begin{minipage}[c]{.32\linewidth}
        \centering
        \resizebox{1.0\linewidth}{!}{
        \begin{tabular}{lccc}
            \toprule
            Method & FVD$_{16}\downarrow$ & KVD$_{16}\downarrow$ \\
            \toprule
            \multicolumn{3}{l}{\textit{Resolution 128$^2$}} \\
            \midrule
            MoCoGAN-HD~\cite{mocogan-hd} & 144.7 $\pm$ 6.0 & 25.4 $\pm$ 1.9 \\
            DIGAN~\cite{digan} & 128.1 $\pm$ 4.9 & 20.6 $\pm$ 1.1 \\
            TATS~\cite{tats} & \textbf{94.6 $\pm$ 2.7} & \textbf{9.8 $\pm$ 1.0} \\
            \midrule
            \multicolumn{3}{l}{\textit{Resolution 256$^2$}} \\
            \midrule
            DIGAN~\cite{digan} & 156.7 $\pm$ 6.2 & - \\
            % TATS~\cite{tats} & \todo{} & \todo{} \\
            % \midrule
            \textbf{\modelname (Ours)} & \textbf{99.0 $\pm$ 2.6} & \textbf{15.3 $\pm$ 0.9}\\
            \bottomrule
        \end{tabular}
        }
    \end{minipage}
    }
    \vspace{-8pt}
    \caption{
    Quantitative comparisons of short video generation on three datasets.
    % Sky Time-lapse~\cite{SkyTimelapse_dataset}, UCF-101~\cite{UCF101}, and TaiChi~\cite{taichi}.
    %
    % \modelname surpasses previous methods under the same resolution by a large margin, while having better performance than state-of-the-art methods under $128^2$ resolution on Sky Time-lapse and UCF-101, and has comparable performance with state-of-the-art methods under $128^2$ resolution on Taichi.
    %
    % \modelname outperforms MCVD on UCF-101 regarding both scores and resolution.
    }
    \label{tab:short-quanti-whole}
\end{table*}

\begin{table}[t] 
    \centering
    \resizebox{1.0\linewidth}{!}{
    \LARGE
    \begin{tabular}{@{}lccccc@{}}
        \toprule
        Method & $\#$Dims & $\#$Params & Step Time & FVD$_{16}\downarrow$ & KVD$_{16}\downarrow$\\
        \toprule
        VDM~\cite{video-dm} & 197k & 445M (397+48) & 4.9s (3.3+1.6) & 1396 & 116 \\ %4.85s (3.25+1.6)
        MCVD~\cite{mcvd} & 786k & 441M & 3s & 2460 & 148 \\
        \midrule
        \textbf{\modelname (Ours)} & 16k & 437M (418+19) & 0.8s & \textbf{552} & \textbf{42}\\
        \bottomrule
        \end{tabular}
    }
    \caption{
        Efficiency and performance comparisons between our latent-space approach with two pixel-space approaches, VDM and MCVD, on unconditional short video generation on UCF-101.
        % (16 frames with a resolution of 256$^2$) on UCF-101.
        %
        All models are trained with approximately the same parameters, the same training time (4.5 days), and the same devices (8 V100s).
        %
        % Batch sizes for all models are 16 except the video super-resolution model of VDM, which we set to 8 (the maximum number under limited GPU memory).
        %
        $+$ indicates two models needed.
        %
        % Our latent space-based method achieves better results in the setting of the same computational resources and the same training time.
        % , demonstrating that our method speeds up the training due to the much more compact latent space.
    }
    \label{tab:efficiency_comparison}
\end{table}
% VDM=64*64*16*3=197k
% MCVD=256*256*4*3=786k
% our=32*32*4*4=

\begin{figure*}[t]
    \centering
    \includegraphics[width=1.0\linewidth]{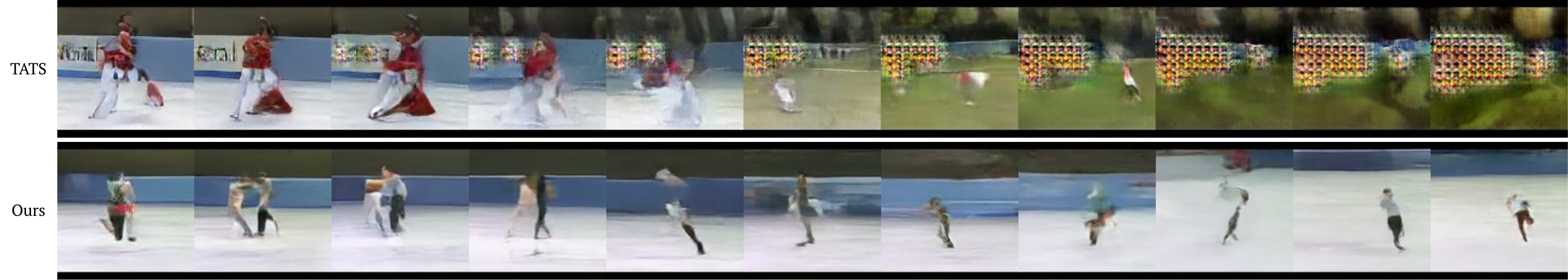}
    \vspace{-20pt}
    \caption{
    Qualitative results of generated long videos compared with state-of-the-art approach TATS\cite{tats} on UCF-101.
    Each frame is selected with a frame step 16.
    Both approaches are compared in the autoregressive setting.
    }
    \label{fig:long-quali}
    \vspace{-0pt}
\end{figure*}

% ----------------------------------------------------------------
% ----------------------------------------------------------------
\begin{figure}[t]
    \centering
    \includegraphics[width=1.0\linewidth]{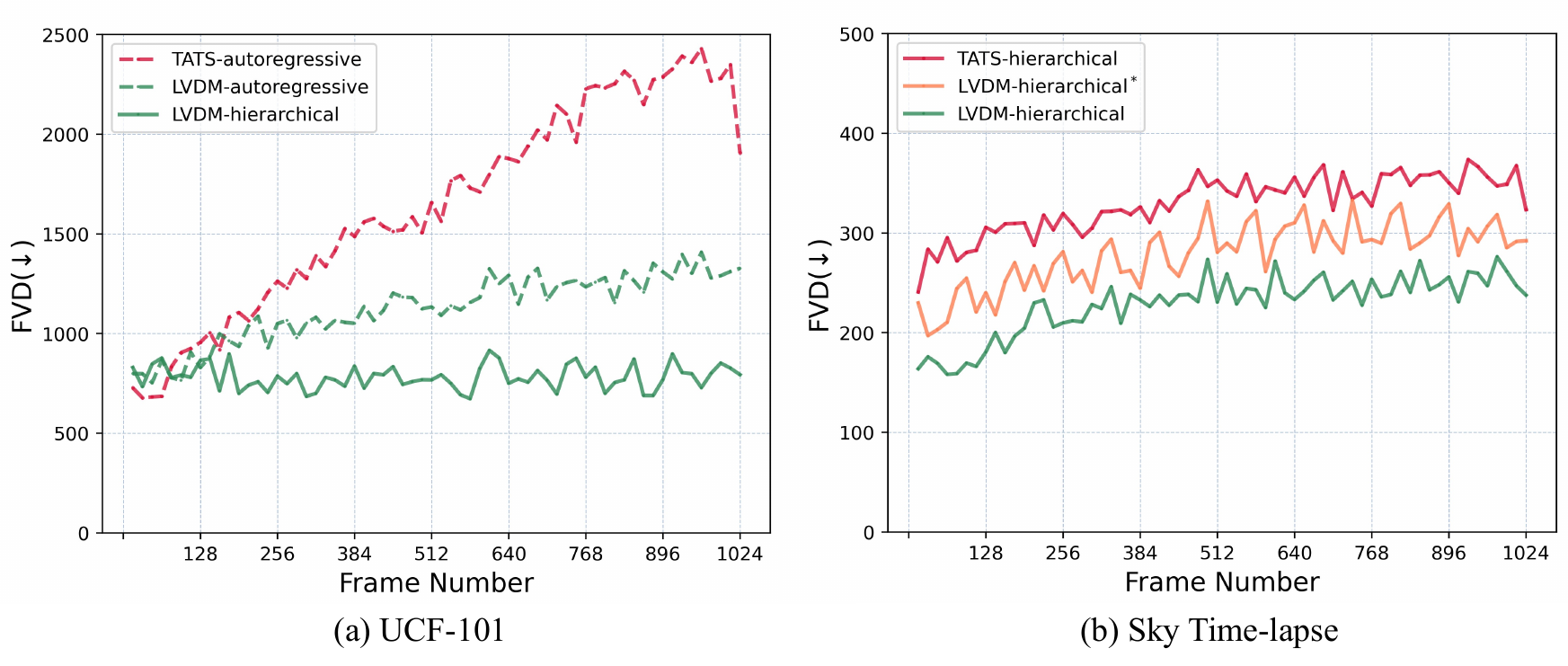}
    \vspace{-20pt}
    \caption{
    Quantitative comparison with TATS on long video generation (1024 frames) on two datasets.
    \modelname outperforms TATS in both autoregressive and hierarchical manners.
    $^*$ indicates DDIM sampling with 100 timesteps.
    }
    \label{fig:long-quanti}
    \vspace{-8pt}
\end{figure}

\subsection{Experimental Setup}
\noindent\textbf{Datasets.} We evaluate our method on UCF-101~\cite{UCF101}, Sky Time-lapse~\cite{SkyTimelapse_dataset}, and Taichi~\cite{taichi}. 
We train all our models \zy{with the resolution of $256^2$ on}
% on 256 resolution 
these datasets for unconditional video generation.
The short clips used for training are selected with \zy{the} frame stride \zy{of} 1 at a random location of one video.
For taichi, we select clips with  \zy{the} frame stride \zy{of} 4 (i.e., skip three frames after selecting one) following prior work~\cite{tats, digan} to make the human motion more dynamic.
Due to the limited number of videos in UCF-101 and TaiChi, we adopt the full dataset for training.
For the Sky Time-lapse dataset, we only \zy{train the model} on its training \zy{split}.
All models are trained under the unconditional setting with no guidance information provided, such as class \zy{label}.

\noindent\textbf{Evaluation Metrics.}
For quantitative evaluation, we report the commonly-used FVD~\cite{FVD} and KVD~\cite{kvd} for both short and long video generation.
Specifically, we calculate FVD and KVD between 2048 real and fake videos \zy{with} 16 frames, which we refer to as FVD$_{16}$ and KVD$_{16}$, for short video evaluation.
All results for short video generation are calculated among ten runs and report their mean and standard deviation.
For long video evaluation, we estimate FVD and KVD among 512 samples in every non-overlapped 16-frame clip and report an FVD  curve across 1024 frames calculated in 1 run, referred to as FVD$_{1024}$.

\noindent\textbf{Baselines.}
We compared our \zy{approach} with seven \zy{competing} baselines, including GAN-based methods: TGAN-v2~\cite{tgan-v2}, DiGAN~\cite{digan}, MoCoGAN-HD~\cite{mocogan-hd}, and long-video GAN~\cite{long-video-gan}, autoregressive models TATS~\cite{tats}, and most \zy{recent} diffusion-based models including Video Diffusion Models (VDM)~\cite{video-dm} and MCVD~\cite{mcvd}.
% StyleGAN-V~\cite{Stylegan-V}, 
Results are collected from their original papers except for VDM.
% or sampled by the provided checkpoints from their official GitHub repo.
%
Please note that VDM has mentioned in its main paper that it will not publish the source code. Thus we implement \zy{it} by ourselves.
Detailed implementations are documented in the supplement.

\noindent\textbf{Implementation Details.}
We use the spatial and temporal downsampling factors of 8 and 4, respectively. We first train a 3D autoencoder, then fix its weights, and then start to train an unconditional \modelname on short video clips.
After that, the \modelname-prediction and \modelname-interpolation models are resumed from the unconditional one.
More details are illustrated in supplementary materials.
% \textbf{Implementation Details.}
% 
% We train the 3D autoencoder with the spatial and temporal downsampling factors \zy{of} 8 and 4, respectively. 
% %
% The channel dimension of latent space is 4, which means a 256 $\times$ 256 $\times$ 16 $\times$ 3 video sample is encoded to \zy{a} 32 $\times$ 32 $\times$ 4 $\times$ 4 latent.
% %
% After the training of autoencoder, we fix its weights and then start to train an unconditional \modelname for short video generation.
% %
% After that, the \modelname-prediction and \modelname-interpolation models are resumed from the unconditional \zy{one}, starting \zy{the joint training} with unconditional-conditional inputs, where the probabilities of unconditional and conditional inputs for the prediction model and interpolation model are (0.5, 0.5) and (0.1, 0.9), respectively.
% %
% When synthesizing long videos, we use unconditional guidance with a scale equal to 0.1,
% %
% and the noise level for conditional frames is set at 200 timesteps.
% %
% Sampling is performed via DDPM standard denoising process unless otherwise specified.
% %
% Most of our models are trained on 8 or 32 A100 GPUs.

% \textbf{Autoencoder}

% \textbf{\modelname-Prediction}

% \textbf{\modelname-Interpolation}

% -------------------------------------------------------------------------------
% first some key points here...
% -------------------------------------------------------------------------------
\subsection{Efficiency Comparison}
To demonstrate the training and sampling efficiency of our method, we compare our approach with two pixel-space video diffusion models, including VDM~\cite{video-dm} and MCVD~\cite{mcvd} in \cref{tab:efficiency_comparison} on the UCF-101 dataset.
We implemented VDM with a base unconditional video diffusion model to synthesize videos of low resolution (16 frames with the resolution of 64$^2$) and a video super-resolution diffusion model to upscale the spatial resolution to 256$^2$.
We train MCVD following its official setting, except that we scale it to resolution 256$^2$.
%
% Specifically, we trained a base unconditional generation video diffusion model to generate 64 $\times$ 64 $\times$ 16 videos and a video super-resolution diffusion model to upscale samples to 256 $\times$ 256 $\times$ 16 high-resolution ones.
%
% We fail to reproduce its gradient-based spatial and temporal extension method. Thus we exploit another common-used design choice method for upscaling the generated samples~\cite{imagen-video}, i.e., train another video super-resolution diffusion model.
%
% MCVD is the most recent work that presents a mask conditional video diffusion model, a joint model for unconditional generation, prediction, and interpolation.
%
%
Our method achieves better FVD \zy{than them} when trained with a similar time and number of model parameters.
%

% -------------------------------------------------------------------------------

\subsection{Short Video Generation}
\noindent\textbf{Quantitative Results.}
In \cref{tab:short-quanti-whole}, we provide quantitative comparisons with previous methods on Sky-Timelapse, UCF-101, and Taichi, respectively.
\zy{Our method outperforms} previous state-of-the-art methods by a large margin.
Specifically, on the Sky-Timelapse dataset, we reduce FVD from 116.5 to 95.18 under 
the resolution \zy{of} 256$^2$. 
In addition, our high-resolution \zy{performance} also surpasses \zy{those} of the state-of-the-art \zy{methods} in FVD under the resolution \zy{of} 128$^2$.
On UCF-101 and taichi datasets, we achieve new state-of-the-art results under the resolution \zy{of} 256$^2$, while our FVD is comparable to the best \zy{result} under the resolution \zy{of} 128$^2$.
\modelname outperforms diffusion-based method MCVD on UCF-101 regarding both FVD and resolution.
%
% Since StyleGAN-V evaluates their approach using a different protocol with TATS, we use their protocol to evaluate ours for a fair comparison.
%
%We achieve 1189.0 on UCF-101, which outperforms StyleGAN-V by a large margin, whose result is 1431.0.
% 1189.0 $\pm$ 24.73
% and Sky Timelapse: \todo{89} FVD$_{16}$

\noindent\textbf{Qualitative Results.}
In \cref{fig:short_video_quali_compare}, we showcase visual comparisons with DIGAN~\cite{digan} and TATS~\cite{tats}.
We \zy{observe} that samples produced by DIGAN exhibit coordinates-like artifacts in many samples, TATS tends to generate samples with flat contents and lacks diversity, while our \modelname can synthesize video samples with high fidelity and diversity.
More comparisons are documented in supplementary materials.

% -------------------------------------------------------------------------------
\subsection{Long Video Generation}
\noindent\textbf{Comparison with State-of-the-art Approach.}
We compare our method \modelname with TATS~\cite{tats} for long video generation with 1024 frames on the UCF-101 and Sky Time-lapse datasets.
Both \modelname and TATS experiment with a pure autoregressive prediction model (autoregressive) and a combination of a prediction model with an interpolation model (hierarchical). Fig.~\ref{fig:long-quali} and Fig.~\ref{fig:long-quanti} present the qualitative and quantitative comparison results, respectively. 
Note that TATS does not provide its hierarchical checkpoints on UCF-101.
%we cannot obtain the official checkpoints of its hierarchical models on UCF-101, we only compare \yty{our \modelname} with its prediction model on this dataset.
On the Sky Time-lapse, we compare two methods under a hierarchical setting.
Our method achieves both lower FVD scores and slower quality degradation over time compared with TATS.
Specifically, on challenging UCF-101, our \modelname-autoregressive has much slower quality degradation over time than TATS-autoregressive.
Our \modelname-hierarchical further significantly alleviates the quality degradation problem.
On Sky Timelapse, both approaches achieve minor quality degradation over time, while our methods have better generation performance.

% -------------------------------------------------------------------------------
% -------------------------------------------------------------------------------

\begin{figure}[t] 
    \centering
    \includegraphics[width=0.99\linewidth]{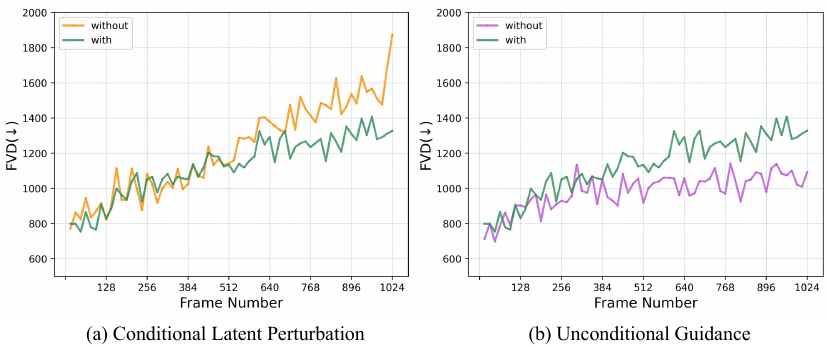}
    \vspace{-8pt}
    \caption{
        Ablation experiments of conditional latent perturbation and unconditional guidance. 
        }
    \label{fig:ablation:uncond_guide}
    \vspace{-0.5cm}
\end{figure}

\noindent\textbf{Conditional Latent Perturbation.}
\cref{fig:ablation:uncond_guide} (a) shows \zy{the} ablation results of conditional latent perturbation.
%
%We set the noise level $s$ for condition is 0 and 200, respectively.
%
We can observe that it can slow down the performance degradation trend, especially when the frame number is larger than 512, demonstrating its effectiveness on long video generation.

\noindent\textbf{Unconditional Guidance.}
We also explore \textit{unconditional guidance} attempting to alleviate quality degradation of autoregressive generation by leveraging the unconditional score to guide the autoregressive conditional generation with a guidance strength. \ytyn{\cref{fig:ablation:uncond_guide} (b) demonstrates that this technique eases the quality degradation of long video generation effectively.} 

\section{Extension for Text-to-Video Generation}
Besides unconditional video generation, we extend our approach to more controllable text-to-video generation.
We scale up our model to a billion-level of parameters and train our model on a 2 million subset of the WebVid dataset.
To efficiently learn video generation models, one natural idea is to leverage the pre-trained text-to-image models to reuse the spatial content generation capacity and continue to learn motion dynamics on the video dataset.
Thus, we initialize the spatial parameters of our model, including the spatial convolutions and spatial attention, using the pre-trained stable diffusion weights.
And we initialize the temporal modules, including the multi-layer temporal self-attention, as identity mapping to preserve the spatial generation performance.
The text-to-video results are shown in \cref{fig:text2video_results}. More video results are documented in supplementary materials.

\section{Conclusion}
In this work, we devise an efficient DM-based framework for video generation, which significantly \zy{reduces} the data dimension and \zy{speeds} up the training and sampling.
Our hierarchical framework can effectively generate long videos with more than one thousand of frames.
% Our novel framework is a joint framework for both unconditional and conditional video synthesis (\ie, prediction and infilling).
%
With its better modeling capacity, we achieve new state-of-the-art results on various datasets under short and long video generation settings.
We further demonstrate the effectiveness of unconditional guidance and conditional latent perturbation in reducing the accumulated error induced during autoregressively extending video lengths.
We also provide an extension of our model for open-domain text-to-video generation.
We hope our method could serve as a strong baseline for future video generation works. Future explorations could be made regarding better architecture design choices and further speeding up the training and sampling of video diffusion models.

%-------------------------------------------------------------------------

{\small
\bibliographystyle{ieee_fullname}
\bibliography{egbib}

\begin{thebibliography}{10}\itemsep=-1pt

\bibitem{biggan}
Andrew Brock, Jeff Donahue, and Karen Simonyan.
\newblock Large scale {GAN} training for high fidelity natural image synthesis.
\newblock In {\em ICLR}, 2019.

\bibitem{long-video-gan}
Tim Brooks, Janne Hellsten, Miika Aittala, Ting-Chun Wang, Timo Aila, Jaakko
  Lehtinen, Ming-Yu Liu, Alexei~A Efros, and Tero Karras.
\newblock Generating long videos of dynamic scenes.
\newblock {\em arXiv preprint arXiv:2206.03429}, 2022.

\bibitem{hierar_video_gen}
Lluis Castrejon, Nicolas Ballas, and Aaron Courville.
\newblock Hierarchical video generation for complex data.
\newblock {\em arXiv preprint arXiv:2106.02719}, 2021.

\bibitem{adm}
Prafulla Dhariwal and Alexander Nichol.
\newblock Diffusion models beat gans on image synthesis.
\newblock {\em Advances in Neural Information Processing Systems},
  34:8780--8794, 2021.

\bibitem{vq-gan}
Patrick Esser, Robin Rombach, and Bjorn Ommer.
\newblock Taming transformers for high-resolution image synthesis.
\newblock In {\em Proceedings of the IEEE/CVF conference on computer vision and
  pattern recognition}, pages 12873--12883, 2021.

\bibitem{tats}
Songwei Ge, Thomas Hayes, Harry Yang, Xi Yin, Guan Pang, David Jacobs, Jia-Bin
  Huang, and Devi Parikh.
\newblock Long video generation with time-agnostic vqgan and time-sensitive
  transformer.
\newblock {\em arXiv preprint arXiv:2204.03638}, 2022.

\bibitem{video-vae}
Jiawei He, Andreas Lehrmann, Joseph Marino, Greg Mori, and Leonid Sigal.
\newblock Probabilistic video generation using holistic attribute control.
\newblock In {\em Proceedings of the European Conference on Computer Vision
  (ECCV)}, pages 452--467, 2018.

\bibitem{imagen-video}
Jonathan Ho, William Chan, Chitwan Saharia, Jay Whang, Ruiqi Gao, Alexey
  Gritsenko, Diederik~P Kingma, Ben Poole, Mohammad Norouzi, David~J Fleet,
  et~al.
\newblock Imagen video: High definition video generation with diffusion models.
\newblock {\em arXiv preprint arXiv:2210.02303}, 2022.

\bibitem{ddpm}
Jonathan Ho, Ajay Jain, and Pieter Abbeel.
\newblock Denoising diffusion probabilistic models.
\newblock {\em Advances in Neural Information Processing Systems},
  33:6840--6851, 2020.

\bibitem{cascaded-dm}
Jonathan Ho, Chitwan Saharia, William Chan, David~J Fleet, Mohammad Norouzi,
  and Tim Salimans.
\newblock Cascaded diffusion models for high fidelity image generation.
\newblock {\em J. Mach. Learn. Res.}, 23:47--1, 2022.

\bibitem{classifier-free}
Jonathan Ho and Tim Salimans.
\newblock Classifier-free diffusion guidance.
\newblock {\em arXiv preprint arXiv:2207.12598}, 2022.

\bibitem{video-dm}
Jonathan Ho, Tim Salimans, Alexey Gritsenko, William Chan, Mohammad Norouzi,
  and David~J Fleet.
\newblock Video diffusion models.
\newblock {\em arXiv preprint arXiv:2204.03458}, 2022.

\bibitem{cogvideo}
Wenyi Hong, Ming Ding, Wendi Zheng, Xinghan Liu, and Jie Tang.
\newblock Cogvideo: Large-scale pretraining for text-to-video generation via
  transformers.
\newblock {\em arXiv preprint arXiv:2205.15868}, 2022.

\bibitem{stylegan3}
Tero Karras, Miika Aittala, Samuli Laine, Erik H\"ark\"onen, Janne Hellsten,
  Jaakko Lehtinen, and Timo Aila.
\newblock Alias-free generative adversarial networks.
\newblock In {\em NeurIPS}, 2021.

\bibitem{stylegan}
Tero Karras, Samuli Laine, and Timo Aila.
\newblock A style-based generator architecture for generative adversarial
  networks.
\newblock In {\em CVPR}, 2019.

\bibitem{stylegan2}
Tero Karras, Samuli Laine, Miika Aittala, Janne Hellsten, Jaakko Lehtinen, and
  Timo Aila.
\newblock Analyzing and improving the image quality of {StyleGAN}.
\newblock In {\em CVPR}, 2020.

\bibitem{videoflow}
Manoj Kumar, Mohammad Babaeizadeh, Dumitru Erhan, Chelsea Finn, Sergey Levine,
  Laurent Dinh, and Durk Kingma.
\newblock Videoflow: A conditional flow-based model for stochastic video
  generation.
\newblock {\em arXiv preprint arXiv:1903.01434}, 2019.

\bibitem{glide}
Alex Nichol, Prafulla Dhariwal, Aditya Ramesh, Pranav Shyam, Pamela Mishkin,
  Bob McGrew, Ilya Sutskever, and Mark Chen.
\newblock Glide: Towards photorealistic image generation and editing with
  text-guided diffusion models.
\newblock {\em arXiv preprint arXiv:2112.10741}, 2021.

\bibitem{im-ddpm}
Alexander~Quinn Nichol and Prafulla Dhariwal.
\newblock Improved denoising diffusion probabilistic models.
\newblock In {\em International Conference on Machine Learning}, pages
  8162--8171. PMLR, 2021.

\bibitem{VQ_VAE}
Aaron van~den Oord, Oriol Vinyals, and Koray Kavukcuoglu.
\newblock Neural discrete representation learning.
\newblock {\em arXiv preprint arXiv:1711.00937}, 2017.

\bibitem{latent-video-transformer}
Ruslan Rakhimov, Denis Volkhonskiy, Alexey Artemov, Denis Zorin, and Evgeny
  Burnaev.
\newblock Latent video transformer.
\newblock {\em arXiv preprint arXiv:2006.10704}, 2020.

\bibitem{dalle-2}
Aditya Ramesh, Prafulla Dhariwal, Alex Nichol, Casey Chu, and Mark Chen.
\newblock Hierarchical text-conditional image generation with clip latents.
\newblock {\em arXiv preprint arXiv:2204.06125}, 2022.

\bibitem{ldm}
Robin Rombach, Andreas Blattmann, Dominik Lorenz, Patrick Esser, and Bj{\"o}rn
  Ommer.
\newblock High-resolution image synthesis with latent diffusion models.
\newblock In {\em Proceedings of the IEEE/CVF Conference on Computer Vision and
  Pattern Recognition}, pages 10684--10695, 2022.

\bibitem{unet}
Olaf Ronneberger, Philipp Fischer, and Thomas Brox.
\newblock U-net: Convolutional networks for biomedical image segmentation.
\newblock In {\em Medical Image Computing and Computer-Assisted
  Intervention--MICCAI 2015: 18th International Conference, Munich, Germany,
  October 5-9, 2015, Proceedings, Part III 18}, pages 234--241. Springer, 2015.

\bibitem{imagen}
Chitwan Saharia, William Chan, Saurabh Saxena, Lala Li, Jay Whang, Emily
  Denton, Seyed Kamyar~Seyed Ghasemipour, Burcu~Karagol Ayan, S~Sara Mahdavi,
  Rapha~Gontijo Lopes, et~al.
\newblock Photorealistic text-to-image diffusion models with deep language
  understanding.
\newblock {\em arXiv preprint arXiv:2205.11487}, 2022.

\bibitem{tgan}
Masaki Saito, Eiichi Matsumoto, and Shunta Saito.
\newblock Temporal generative adversarial nets with singular value clipping.
\newblock In {\em Proceedings of the IEEE international conference on computer
  vision}, pages 2830--2839, 2017.

\bibitem{tgan-v2}
Masaki Saito, Shunta Saito, Masanori Koyama, and Sosuke Kobayashi.
\newblock Train sparsely, generate densely: Memory-efficient unsupervised
  training of high-resolution temporal gan.
\newblock {\em International Journal of Computer Vision}, 128:2586--2606, 2020.

\bibitem{taichi}
Aliaksandr Siarohin, St{\'e}phane Lathuili{\`e}re, Sergey Tulyakov, Elisa
  Ricci, and Nicu Sebe.
\newblock First order motion model for image animation.
\newblock {\em NeurIPS}, 2019.

\bibitem{make-a-video}
Uriel Singer, Adam Polyak, Thomas Hayes, Xi Yin, Jie An, Songyang Zhang, Qiyuan
  Hu, Harry Yang, Oron Ashual, Oran Gafni, et~al.
\newblock Make-a-video: Text-to-video generation without text-video data.
\newblock {\em arXiv preprint arXiv:2209.14792}, 2022.

\bibitem{Stylegan-V}
Ivan Skorokhodov, Sergey Tulyakov, and Mohamed Elhoseiny.
\newblock Stylegan-v: A continuous video generator with the price, image
  quality and perks of stylegan2.
\newblock In {\em Proceedings of the IEEE/CVF Conference on Computer Vision and
  Pattern Recognition}, pages 3626--3636, 2022.

\bibitem{dynamic}
Jascha Sohl-Dickstein, Eric Weiss, Niru Maheswaranathan, and Surya Ganguli.
\newblock Deep unsupervised learning using nonequilibrium thermodynamics.
\newblock In {\em International Conference on Machine Learning}, pages
  2256--2265. PMLR, 2015.

\bibitem{ddim}
Jiaming Song, Chenlin Meng, and Stefano Ermon.
\newblock Denoising diffusion implicit models.
\newblock {\em arXiv preprint arXiv:2010.02502}, 2020.

\bibitem{score-sde}
Yang Song, Jascha Sohl-Dickstein, Diederik~P Kingma, Abhishek Kumar, Stefano
  Ermon, and Ben Poole.
\newblock Score-based generative modeling through stochastic differential
  equations.
\newblock {\em arXiv preprint arXiv:2011.13456}, 2020.

\bibitem{UCF101}
Khurram Soomro, Amir~Roshan Zamir, and Mubarak Shah.
\newblock Ucf101: A dataset of 101 human actions classes from videos in the
  wild.
\newblock {\em arXiv preprint arXiv:1212.0402}, 2012.

\bibitem{mocogan-hd}
Yu Tian, Jian Ren, Menglei Chai, Kyle Olszewski, Xi Peng, Dimitris~N. Metaxas,
  and Sergey Tulyakov.
\newblock A good image generator is what you need for high-resolution video
  synthesis.
\newblock In {\em International Conference on Learning Representations}, 2021.

\bibitem{mocogan}
Sergey Tulyakov, Ming-Yu Liu, Xiaodong Yang, and Jan Kautz.
\newblock Mocogan: Decomposing motion and content for video generation.
\newblock In {\em Proceedings of the IEEE conference on computer vision and
  pattern recognition}, pages 1526--1535, 2018.

\bibitem{FVD}
Thomas Unterthiner, Sjoerd van Steenkiste, Karol Kurach, Raphael Marinier,
  Marcin Michalski, and Sylvain Gelly.
\newblock Towards accurate generative models of video: A new metric \&
  challenges.
\newblock {\em arXiv preprint arXiv:1812.01717}, 2018.

\bibitem{kvd}
Thomas Unterthiner, Sjoerd van Steenkiste, Karol Kurach, Raphael Marinier,
  Marcin Michalski, and Sylvain Gelly.
\newblock Towards accurate generative models of video: A new metric \&
  challenges.
\newblock {\em ICLR}, 2019.

\bibitem{mcvd}
Vikram Voleti, Alexia Jolicoeur-Martineau, and Christopher Pal.
\newblock Masked conditional video diffusion for prediction, generation, and
  interpolation.
\newblock {\em arXiv preprint arXiv:2205.09853}, 2022.

\bibitem{vgan}
Carl Vondrick, Hamed Pirsiavash, and Antonio Torralba.
\newblock Generating videos with scene dynamics.
\newblock {\em Advances in neural information processing systems}, 29, 2016.

\bibitem{vqvae-prediction}
Jacob Walker, Ali Razavi, and A{\"a}ron van~den Oord.
\newblock Predicting video with vqvae.
\newblock {\em arXiv preprint arXiv:2103.01950}, 2021.

\bibitem{scale-autoregressive}
Dirk Weissenborn, Oscar T{\"a}ckstr{\"o}m, and Jakob Uszkoreit.
\newblock Scaling autoregressive video models.
\newblock {\em arXiv preprint arXiv:1906.02634}, 2019.

\bibitem{SkyTimelapse_dataset}
Wei Xiong, Wenhan Luo, Lin Ma, Wei Liu, and Jiebo Luo.
\newblock Learning to generate time-lapse videos using multi-stage dynamic
  generative adversarial networks.
\newblock In {\em The IEEE Conference on Computer Vision and Pattern
  Recognition (CVPR)}, June 2018.

\bibitem{VideoGPT}
Wilson Yan, Yunzhi Zhang, Pieter Abbeel, and Aravind Srinivas.
\newblock Videogpt: Video generation using vq-vae and transformers.
\newblock {\em arXiv preprint arXiv:2104.10157}, 2021.

\bibitem{pvdm}
Sihyun Yu, Kihyuk Sohn, Subin Kim, and Jinwoo Shin.
\newblock Video probabilistic diffusion models in projected latent space.
\newblock {\em arXiv preprint arXiv:2302.07685}, 2023.

\bibitem{digan}
Sihyun Yu, Jihoon Tack, Sangwoo Mo, Hyunsu Kim, Junho Kim, Jung-Woo Ha, and
  Jinwoo Shin.
\newblock Generating videos with dynamics-aware implicit generative adversarial
  networks.
\newblock In {\em International Conference on Learning Representations}, 2022.

\bibitem{lpips}
Richard Zhang, Phillip Isola, Alexei~A Efros, Eli Shechtman, and Oliver Wang.
\newblock The unreasonable effectiveness of deep features as a perceptual
  metric.
\newblock In {\em CVPR}, 2018.

\bibitem{magicvideo}
Daquan Zhou, Weimin Wang, Hanshu Yan, Weiwei Lv, Yizhe Zhu, and Jiashi Feng.
\newblock Magicvideo: Efficient video generation with latent diffusion models.
\newblock {\em arXiv preprint arXiv:2211.11018}, 2022.

\end{thebibliography}
}

\end{document}